\pdfoutput=1

\documentclass[11pt]{article}

\usepackage[final]{acl}

\usepackage{times}
\usepackage{latexsym}

\usepackage[T1]{fontenc}

\usepackage[utf8]{inputenc}

\usepackage{microtype}
\usepackage{subcaption}
\usepackage{inconsolata}

\usepackage{graphicx}

\usepackage{booktabs}       
\usepackage{amsfonts}       
\usepackage{nicefrac}       
\usepackage{microtype}      
\usepackage[table, dvipsnames]{xcolor}         
\usepackage{graphicx}
\usepackage{adjustbox}
\usepackage{multirow}
\usepackage{wrapfig}
\usepackage{amsmath}
\usepackage{enumitem}
\usepackage{makeidx} 
\usepackage{appendix} 
\usepackage{titletoc}
\usepackage{floatflt}
\usepackage{ragged2e}
\usepackage[most]{tcolorbox} 
\usepackage{tabularx}
\usepackage{makecell}
\usepackage{svg}
\usepackage{soul,xcolor}
\soulregister\cite7

\tcbuselibrary{listings, breakable}
\usepackage[dvipsnames]{xcolor}
\usepackage{listings}
\usepackage{fancyvrb}
\usepackage{float}
\usepackage{placeins}
\usepackage{caption} 
\fvset{
    formatcom=\color{py-doc},
}
\lstdefinestyle{pydoc}{
    basicstyle=\ttfamily\color{py-doc},
    breaklines=true,
    breakatwhitespace=false,
    xleftmargin=2em
}

\usepackage{times}
\usepackage{latexsym}

\usepackage[T1]{fontenc}

\usepackage[utf8]{inputenc}

\usepackage{microtype}

\usepackage{inconsolata}

\usepackage{graphicx}

\usepackage{multirow}
\usepackage{array}
\usepackage{color}
\usepackage{tabularx}
\newcolumntype{Y}{>{\centering\arraybackslash}X}
\usepackage{wrapfig}
\usepackage{caption}
\usepackage{booktabs}
\usepackage{comment}
\usepackage{float}
\usepackage{stfloats}
\usepackage{subcaption}
\usepackage{algorithm, algorithmic}

\usepackage{booktabs}
\usepackage[table]{xcolor}
\usepackage{graphicx}
\usepackage{adjustbox}
\usepackage{calc}
\definecolor{cxrule}{gray}{0.8}

\usepackage{lipsum}

\newcolumntype{C}[1]{>{\centering\arraybackslash}p{#1}} 
\newcolumntype{M}[1]{>{\centering\arraybackslash}m{#1}}

\title{ChartEvo: Self-Evolving Agent Harnesses for Long-Tail Chart Reasoning}

\author{
    Qi Wang$^{1,2}$\thanks{~~~Equal contribution.},
    Peijie Wang$^{1,2}$\footnotemark[1],
    Jun Cao$^{1,2}$,
    Haoyang Tong$^{1,3,4}$\\
    \textbf{Xiao-Hui Li$^{1,2}$,
    Fei Yin$^{1,2}$,
    Cheng-Lin Liu$^{1,2}$}\thanks{~~~Corresponding author.}\\
    $^1$State Key Laboratory of Multimodal Artificial Intelligence Systems, CASIA\\
    $^2$School of Artificial Intelligence, University of Chinese Academy of Sciences\\
    $^3$New Laboratory of Pattern Recognition, CASIA\\
    $^4$Zhongguancun Academy\\
    \texttt{wangqi2026@ia.ac.cn \quad wangpeijie2023@ia.ac.cn} \\
    \texttt{\{fyin, liucl\}@nlpr.ia.ac.cn}\\
    \rule{0pt}{1.5cm}
}

\colorlet{pale1}{blue!10}
\colorlet{pale2}{green!10}
\colorlet{pale3}{red!10}
\colorlet{pale4}{orange!10}
\colorlet{pale5}{cyan!10}
\colorlet{pale6}{magenta!10}
\colorlet{pale7}{gray!10}
\colorlet{pale8}{teal!10}
\colorlet{pale9}{purple!10}
\newcolumntype{P}[1]{>{\centering\arraybackslash}p{#1}}
\usepackage{pifont}

\begin{document}
\maketitle

\begin{abstract}
Multimodal large language models (MLLMs) have substantially advanced chart understanding and reasoning, yet their performance remains highly uneven across chart types. While modern MLLMs perform well on common charts, their accuracy drops sharply on long-tail charts with diverse visual structures and reasoning requirements. Tool-augmented agents can improve complex chart reasoning by introducing specialized visual capabilities, but manually engineering strategies and tools for the heterogeneous and evolving long tail is difficult to scale. We introduce \textbf{ChartEvo}, a recursive self-improvement framework that enables chart agents to adapt to underperforming chart types through experience-driven harness-level self-improvement. ChartEvo jointly evolves reusable chart-specific Skills and executable Tools, while learning reusable local tool-call chains to compose evolved capabilities and control the inference overhead of tool-augmented reasoning. On long-tail charts from ChartBench and ChartX, ChartEvo improves accuracy over its base MLLM, Qwen3.7-Plus, by \textbf{45.6\%} and \textbf{32.5\%}, respectively, demonstrating the effectiveness of recursive self-improvement for long-tail chart adaptation.
\end{abstract}

\section{Introduction}

\begin{figure}[h]
\vspace{-0.4cm}
    \centering
    \includegraphics[width=1.0\linewidth]{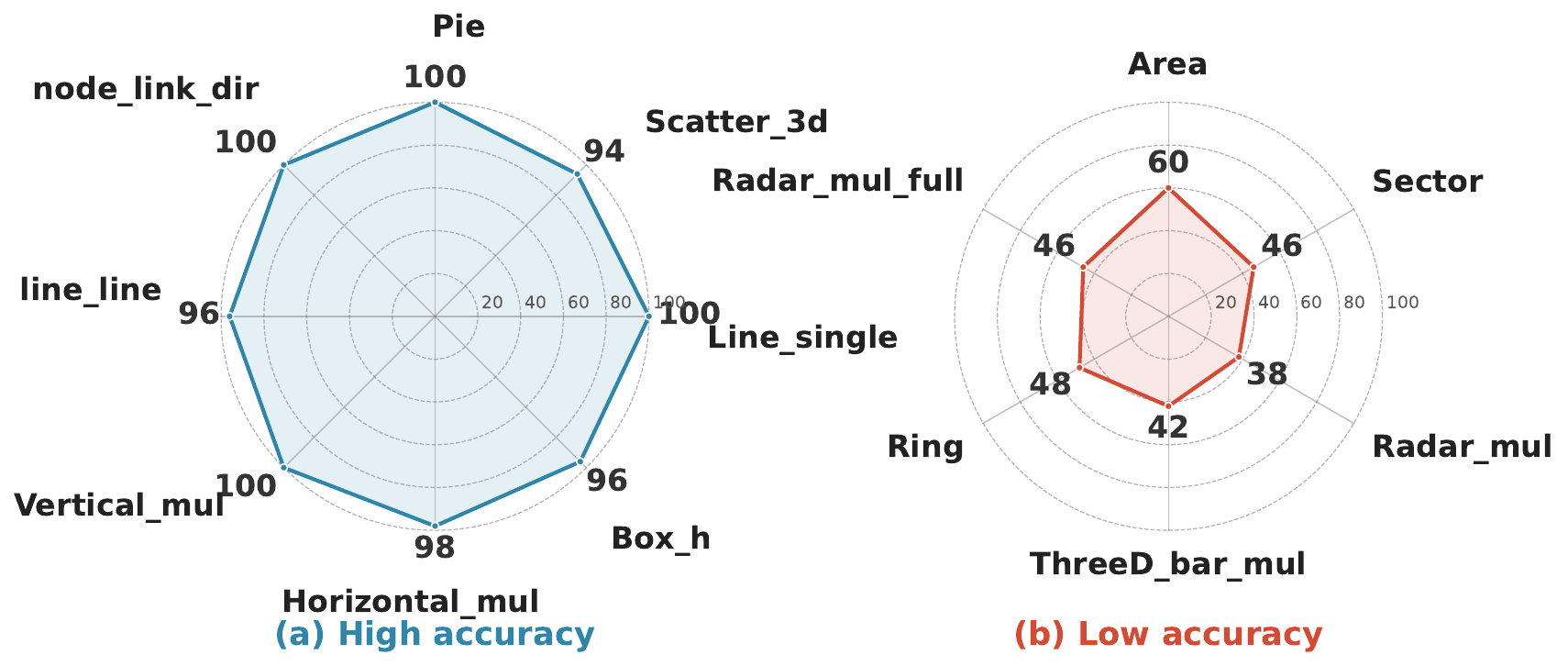}
    \vspace{-0.6cm}
    \caption{Representative well-performing and underperforming chart types in ChartBench test split.}
    \label{fig:question_explain}
\vspace{-0.3cm}
\end{figure}

Recent advances in multimodal large language models (MLLMs) have substantially improved chart understanding and reasoning, yielding strong overall performance on established chart question answering (CQA) benchmarks~\cite{chartqa,chartbench}. Tool-augmented approaches such as ChartAgent~\cite{chartagent} further extend chart reasoning by incorporating specialized visual tools for complex visual and numerical tasks. However, strong aggregate performance can obscure substantial performance disparities across chart types~\cite{chartqapro, chartmind}. On the numerical question answering (NQA) task in ChartBench, Qwen3.7-Plus achieves an average accuracy of 80.0\% across 26 chart types. Among them, 17 types exceed 90\% accuracy, with an average of 96.9\%, whereas the remaining nine average only 48.0\%. As illustrated in Fig.~\ref{fig:question_explain}, this disparity reveals a highly uneven performance distribution: while many chart types are handled reliably, the remaining underperforming types form a heterogeneous tail with distinct visual structures and reasoning requirements. We refer to these persistently underperforming chart types as \emph{long-tail charts}. This uneven performance distribution suggests that further progress in CQA increasingly depends on an agent's ability to adapt to these diverse, chart-specific capability gaps.

\begin{figure*}[t]
\vspace{-0.2cm}
    \centering
    \includegraphics[width=1.0\linewidth]{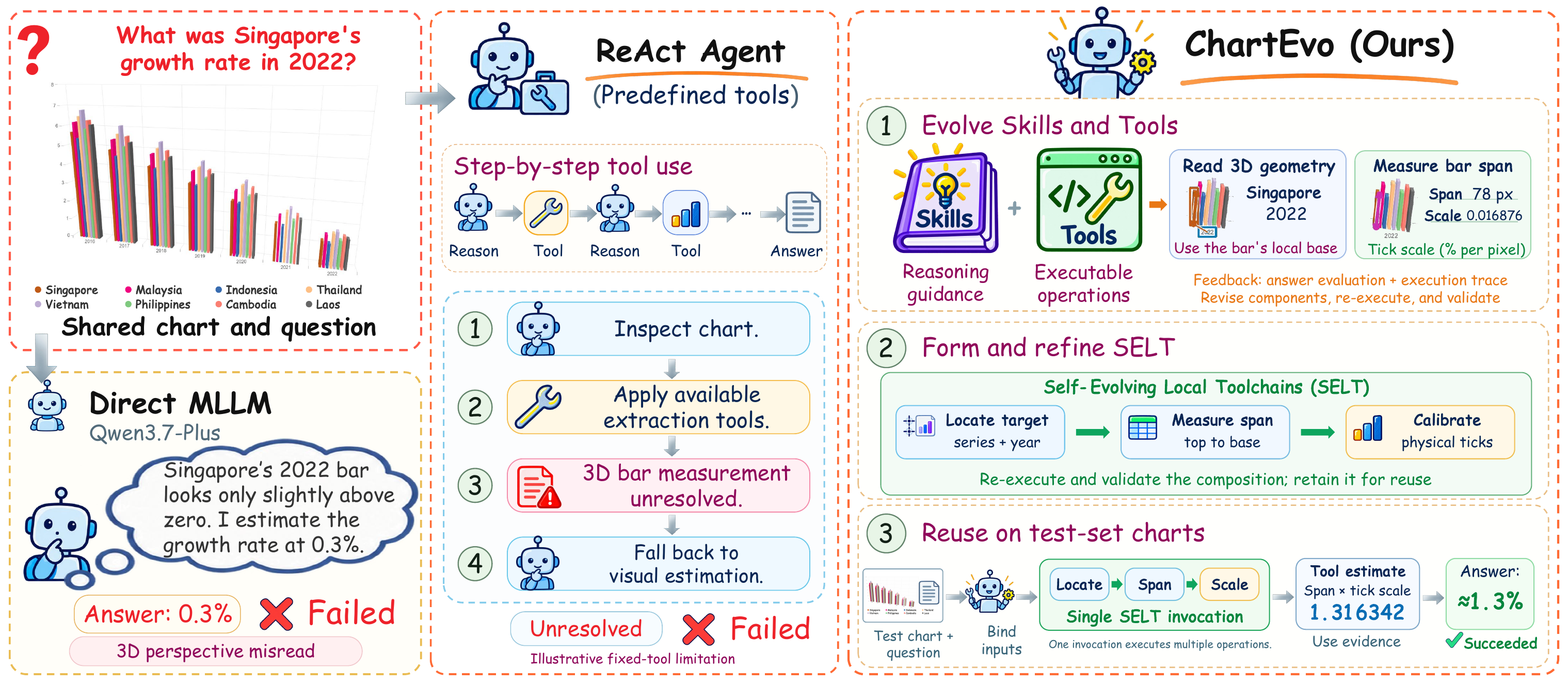}
    \vspace{-0.6cm}
    \caption{Illustrative comparison of chart reasoning approaches. Direct MLLMs and fixed-tool agents can misinterpret chart encodings. ChartEvo evolves skills and tools into validated, reusable Self-Evolving Local Toolchains (SELT), enabling evidence-grounded reasoning through a single toolchain invocation.}
    \label{fig:intro}
\vspace{-0.4cm}
\end{figure*}

Addressing this heterogeneous long tail requires capable MLLMs to continually acquire new capabilities for chart types they have not yet mastered. One possible route is to further train the model, encoding the missing capabilities into its parameters. However, repeatedly adapting model parameters to address localized capability gaps across diverse chart structures and reasoning patterns is neither naturally modular nor readily extensible. Tool-augmented chart agents provide a more modular alternative by introducing specialized visual operations that complement the MLLM. Yet, existing approaches largely rely on manually predefined tools and usage strategies, leaving the agent's capability boundary constrained by human engineering. As unresolved failures span diverse chart structures and reasoning requirements, manually extending the agent for each distinct failure mode becomes difficult to scale. This motivates shifting long-tail adaptation from manual capability engineering toward \emph{autonomous capability acquisition}.

To this end, we introduce \textbf{ChartEvo}, a recursively self-improving chart agent that autonomously  evolves its harness to acquire and organize capabilities for long-tail chart reasoning. An illustrative overview of ChartEvo and its contrast with direct MLLM and fixed-tool reasoning is shown in Fig.~\ref{fig:intro}. ChartEvo progressively accumulates reusable capabilities in its external harness. Specifically, it jointly evolves chart-specific Skills that capture reusable reasoning knowledge and executable Tools that expand the agent's visual and numerical capabilities. Skills thus encode how to reason, while Tools extend what the agent can execute, enabling acquired capabilities to be retained and effectively applied across related chart instances.

Beyond capability acquisition, a complementary challenge lies in how to efficiently organize and execute these capabilities. Conventional ReAct-style agents repeatedly interleave reasoning, tool invocation, and observation~\cite{react}. In chart reasoning, many tasks require global reasoning to determine what should be measured, but once the target and procedure are established, the remaining computation often consists of locally structured visual operations with stable dependencies. Executing each intermediate step through the ReAct loop therefore requires routing tool outputs back through the MLLM, introducing additional model--tool interaction overhead. We introduce \textbf{Self-Evolving Local Toolchains (SELT)} to address this issue. Through SELT, ChartEvo evolves and preserves validated local Tool compositions as reusable execution structures, enabling structured procedures to execute with fewer MLLM-mediated tool interactions while retaining the flexibility of dynamic Tool composition for previously unseen cases. Skills capture reusable reasoning knowledge, Tools expand executable capabilities, and SELT organize how these capabilities are locally composed and executed. ChartEvo therefore evolves not only the agent's capability boundary but also the execution structure through which these capabilities are applied. Comprehensive experiments validate the effectiveness of ChartEvo for long-tail chart reasoning and its ability to control the execution overhead of tool-augmented reasoning.
Our contributions are summarized as follows:

\noindent \textbf{1. Harness Recursive Self-Improvement:}
We introduce \textbf{ChartEvo}, a harness-level recursive self-improvement framework that autonomously evolves reusable Skills and executable Tools through iterative feedback, enabling continual capability acquisition for long-tail chart reasoning.

\noindent \textbf{2. Self-Evolving Local Toolchains:}
We propose \textbf{SELT}, extending harness-level self-improvement to execution structures by evolving and reusing validated Tool compositions, thereby reducing redundant model--tool interactions and improving execution efficiency in ReAct-style reasoning.

\noindent \textbf{3. Effective and Efficient Long-Tail Chart Reasoning:} On long-tail charts from ChartBench and ChartX, ChartEvo improves accuracy over the baseline MLLM by \textbf{45.6\%} and \textbf{32.5\%}, respectively, while substantially reducing model calls with comparable accuracy.

\vspace{-0.2cm}
\section{Related Work}
\vspace{-0.2cm}

\subsection{Chart Question Answering}
\vspace{-0.1cm}
Early CQA studies formulated chart understanding as task-specific visual question answering, progressing from synthetic charts and template-based questions~\cite{figureqa, DVQA} to real-world charts and more complex numerical and logical reasoning~\cite{PlotQA, LEAFQA, chartqa}. Vision-language pretraining further enabled more general chart understanding through derendering and structured representations~\cite{matcha,unichart}.

With the rise of MLLMs, recent studies have adapted general-purpose models to CQA through chart-specific instruction tuning and multimodal alignment~\cite{chartllama,chartinstruct,chartassisstant}, while increasingly challenging benchmarks expose persistent limitations in precise visual and numerical reasoning~\cite{chartbench,chartx,CharXiv,chartqapro}. Most recently, ChartAgent addresses such limitations through tool-augmented visual reasoning~\cite{chartagent}. However, its reliance on predefined tools and usage strategies makes manual capability engineering difficult to scale across long-tail charts with diverse structures and reasoning requirements, motivating us to explore RSI for autonomously expanding agent capabilities.

\vspace{-0.2cm}
\subsection{Recursive Self‑Improvement}
\vspace{-0.1cm}
Recursive self-improvement (RSI) studies how AI systems leverage self-generated experience to improve their capabilities. Existing research explores complementary pathways, including updating model parameters through self-generated data and reinforcement learning~\cite{spin,lsp,spell}, as well as evolving the external systems that mediate agent behavior. We focus on the latter. Gödel Agent enables agents to recursively modify their logic and behavior, while Meta-Harness further treats the entire agent harness as an optimization object, extending self-improvement to the system level~\cite{godel,metaharness}.

Recent work increasingly instantiates such improvement through modular and reusable agent capabilities~\cite{skillsmith, coevoskill, anything2skill, managing}. SkillForge, SkillOpt, and MetaSkill-Evolve enable agents to construct and refine reusable skills from execution experience~\cite{skillforge,skillopt,metaskill}. In parallel, early studies such as CREATOR and LATM established autonomous tool creation~\cite{creator,latm}, while recent self-evolving agents increasingly incorporate autonomous tool discovery, construction, repair, and maintenance~\cite{toolmaker,stella,toolmaking, toolscope}. Building on these advances, we jointly evolve chart-specific Skills and Tools for long-tail chart reasoning, while learning reusable local tool-call chains for efficient execution.

\vspace{-0.2cm}
\section{Methodology}
\vspace{-0.2cm}

\subsection{Overview}
\vspace{-0.1cm}

ChartEvo treats the external harness of a chart-reasoning agent as the object of continual self-evolution. Given a chart image $I$ and a question $q$, the underlying model reasons with the current harness to produce an answer. In ChartEvo, the harness consists of Skills, Tools, and Self-Evolving Local Toolchains (SELT). Skills encode reusable chart-specific reasoning and capability-use knowledge, Tools provide executable visual and numerical operations, and SELT persist validated compositions of Tool calls together with their execution dependencies. Skills and Tools thus define the reasoning knowledge and executable capabilities available to the agent, while SELT organize how these capabilities are composed and executed.

\begin{figure*}[t]
\vspace{-0.2cm}
    \centering
    \includegraphics[width=1.0\linewidth]{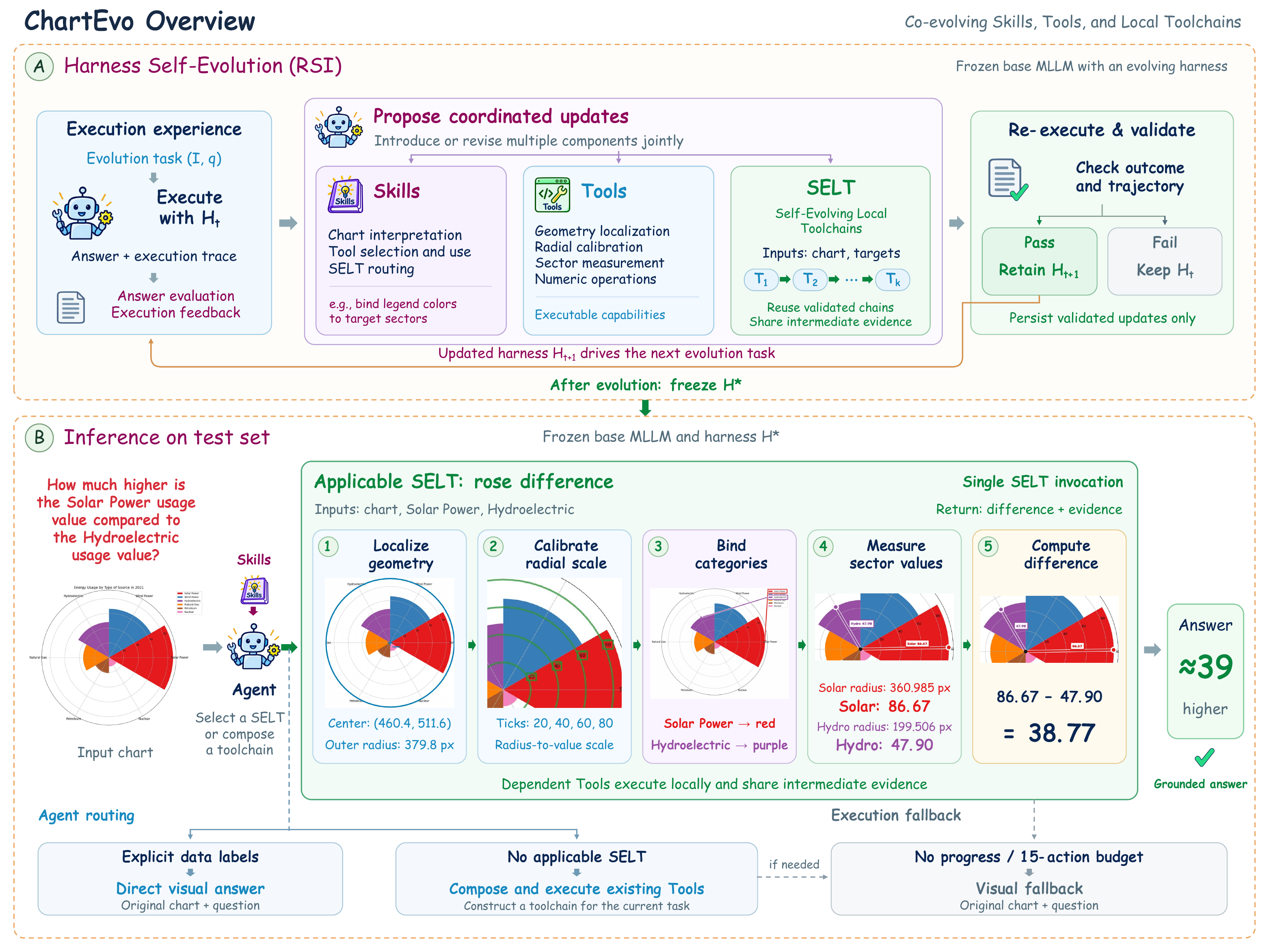}
    \vspace{-8mm}
    \caption{Overview of ChartEvo. (A) Harness self-evolution jointly updates Skills, Tools, and SELT using answer evaluation and execution feedback, retaining validated updates for subsequent rounds. (B) With the harness frozen at test time, the rose-chart example illustrates five local reasoning stages within a single SELT invocation.}
    \label{fig:pipeline}
\vspace{-0.6cm}
\end{figure*}

As illustrated in Fig.~\ref{fig:pipeline}, ChartEvo formulates harness-level recursive self-improvement as an iterative outer-loop process. Given the current harness and its execution outcomes on chart-reasoning tasks, an agentic proposer introduces or revises Skills, Tools, and SELT. Validated updates are persisted into the next harness state, which is subsequently reused for chart reasoning and serves as the basis for further harness evolution.

During self-evolution, supervision signals are used to evaluate execution outcomes and guide harness updates. After evolution, the resulting harness is deployed for inference on unseen chart instances. Overall, ChartEvo instantiates the agent as a \emph{Model + Evolving Harness} system: the model serves as the general reasoning core, while the harness progressively accumulates Skills, Tools, and SELT to acquire new capabilities and more structured execution patterns for long-tail chart reasoning.

\vspace{-0.2cm}
\subsection{Harness Recursive Self-Improvement}
\vspace{-0.1cm}

ChartEvo starts self-evolution from initial harness $\mathcal{H}_0$ containing a set of generic chart-processing Tools. These Tools provide the basic primitives for bootstrapping chart reasoning, while task-specific capabilities are progressively acquired through self-evolution. At evolution step $t$, the current harness $\mathcal{H}_t$ is used to solve a task, producing both a prediction and its execution trajectory $\tau_t$. The agentic proposer jointly considers the final answer, task-level evaluation feedback, and the observed execution trajectory to analyze the behavior of the current harness and identify components that require further improvement. This joint observation allows the proposer to assess both task-level correctness and the reliability of the execution process, including cases where an apparently acceptable prediction is supported by an unreliable trajectory.

Based on the current harness, its execution trajectory, and the supervision available during self-evolution, the agentic proposer can introduce new harness components or revise existing Skills, Tools, and SELT. ChartEvo allows multiple component types to evolve jointly within the same step, enabling coordinated updates to reasoning knowledge, executable capabilities, and execution structures. A proposal may therefore span multiple harness components and be evaluated as a coordinated update to the current harness. Ground-truth answers are available during self-evolution for task evaluation and candidate validation, while no ground-truth solution traces are provided.

Candidate updates are validated by re-executing the task under the updated harness. Both the resulting task outcome and the corresponding execution trajectory are examined to determine whether the updated harness behaves as intended. Validated components and structures are then persisted into $\mathcal{H}_{t+1}$, which becomes the basis for subsequent execution and further self-evolution. Through repeated execution, analysis, proposal or revision, and validation, ChartEvo recursively evolves its harness over successive tasks, such that subsequent proposals are made over an already evolved harness.

Sec.~\ref{sec3.3}, Sec.~\ref{sec3.4}, and Sec.~\ref{sec3.5} describe Skill/Tool evolution, Self-Evolving Local Toolchains, and inference with the evolved harness, respectively.

\vspace{-0.2cm}
\subsection{Skill and Tool Evolution}
\label{sec3.3}
\vspace{-0.1cm}

ChartEvo treats Skills and Tools as two complementary carriers of harness capabilities. Skills encode reusable reasoning and capability-use knowledge, guiding the model in how to interpret, select, and use available capabilities. Tools provide executable visual and numerical operations that transform chart information into structured evidence consumable by subsequent reasoning. Together, they define the reasoning knowledge and executable capabilities available to the current harness.
During self-evolution, agentic proposer determines whether an observed deficiency primarily arises from how existing capabilities are composed and used, or from the lack of an executable evidence capability. When the available Tools can already provide the required information but their selection, invocation, or use does not support correct reasoning, evolution primarily targets Skills. When the existing Tool set cannot provide the required structured evidence, or when a Tool exhibits incorrect behavior, insufficient coverage, or an inadequate output interface, the proposer introduces or revises Tools. Existing components are revised when their intended semantics remain valid, whereas new components are introduced when the required reasoning pattern or executable capability is semantically distinct from those already available.

Skill and Tool evolution can be coupled within the same self-evolution step, allowing newly introduced or revised executable capabilities to be incorporated into corresponding reasoning and capability-use knowledge. Candidate Skills and Tools are re-executed and validated under the updated harness, with Tools additionally checked for executable behavior. Validated updates are persisted in the harness and remain available for reasoning and further evolution. While Skill and Tool evolution expands the capabilities available to the harness, stable multi-step execution structures over these capabilities are captured through SELT.

\vspace{-0.2cm}
\subsection{Self-Evolving Local Toolchains}
\label{sec3.4}
\vspace{-0.1cm}

Self-Evolving Local Toolchains (SELT) persist validated Tool compositions as reusable execution structures, avoiding repeated reconstruction of the same Tool-call patterns through model-mediated interaction. Each SELT represents a bounded executable transformation with explicit inputs, outputs, execution order, and internal Tool dependencies. The boundedness of a SELT is defined by its functional contract: the Agent resolves the task-specific target and required arguments, after which the SELT executes the structured transformation and returns its designated output.

During self-evolution, the Agent can reuse an existing SELT when an applicable execution structure is available. Otherwise, it dynamically composes individual Tools, coordinating their execution and intermediate results through model interactions. Validated compositions with stable execution structures can subsequently be persisted as new SELT, while existing SELT can be revised as Tools, Skills, or execution requirements evolve. Unlike dynamic Tool composition, SELT delegate inter-step parameter resolution and intermediate output propagation to the runtime, eliminating repeated Agent planning and binding within established Toolchains.

SELT usage is jointly mediated by Skills, the Agent, and the runtime. Skills maintain the available SELT together with their routing and usage knowledge. The runtime progressively discloses context-relevant candidate SELT according to the current question, execution stage, and accumulated observations, exposing only a relevant subset of the harness capabilities at each decision step. The Agent selects a concrete SELT based on question semantics, available evidence, and Skill-provided routing knowledge, and binds the task-specific arguments required for execution. The runtime then executes the selected SELT while enforcing invocation constraints, Tool versions, and execution checks.

Skills and SELT therefore play complementary roles: Skills specify when and how an execution structure should be used, whereas SELT encode its concrete Tool composition and dependencies. Once the Agent binds the task target, entities, and execution conditions, the selected SELT executes the structured procedure without further model-mediated coordination of internal Tool calls. When no applicable SELT is available, the Agent can fall back to dynamic Tool composition, whose successful execution may drive subsequent SELT evolution.

\vspace{-0.2cm}
\subsection{Inference with the Evolved Harness}
\label{sec3.5}
\vspace{-0.1cm}

After self-evolution, ChartEvo fixes the resulting harness and applies it to unseen chart instances. The Agent uses the evolved Skills, Tools, and SELT according to the current question and execution context. When an applicable SELT is available, its persistent execution structure can be directly reused; otherwise, the Agent can dynamically compose Tools to complete the task. To handle unsuccessful execution, ChartEvo supports two fallback conditions: the Agent may proactively switch to direct visual answering when it determines that harness-assisted reasoning can no longer proceed, or the system triggers fallback when the maximum action budget of 15 is reached. In both cases, direct visual answering is performed using only the original chart image and question.

\begin{table*}[t]
\centering
\caption{\textbf{Comparison on ChartBench.}
NQA accuracy (\%) on nine long-tail chart types.
The best and second-best results are highlighted in
\textbf{bold} and \underline{underline}, respectively.}
\label{tab:chartbench}
\vspace{-0.2cm}

\begingroup
\scriptsize
\setlength{\tabcolsep}{4pt}
\renewcommand{\arraystretch}{1.18}

\definecolor{cbheader}{HTML}{F5F3F1}
\definecolor{cbstripe}{HTML}{FAF9F8}
\definecolor{cbours}{HTML}{FDEBDC}
\definecolor{cborange}{HTML}{E96B25}
\definecolor{cbrule}{HTML}{BDB7B0}
\definecolor{cbgrouptext}{HTML}{655F59}

\arrayrulecolor{cbrule}

\newcommand{\cbbest}[1]{\textbf{#1}}
\newcommand{\cbsecond}[1]{\underline{#1}}

\newcommand{\cbgroup}[1]{%
  \multicolumn{11}{c}{%
    \color{cbgrouptext}\bfseries\itshape #1%
  }\\
}

\resizebox{\textwidth}{!}{%
\begin{tabular}{l|c*{8}{c}|c}
\toprule[1.0pt]

\rowcolor{cbheader}
\textbf{Method}
& \textbf{Area}
& \textbf{3Dbar-multi}
& \textbf{3D-stack}
& \textbf{Ring}
& \textbf{Sector}
& \textbf{Radar-M}
& \textbf{Radar-MF}
& \textbf{Radar-S}
& \textbf{Stock}
& \textbf{Avg.} \\
\midrule[0.6pt]

\rowcolor{cbheader}
\cbgroup{Direct Visual MLLMs}

DeepSeek-v4.1-Flash
& 30.0 & 38.0 & 12.0 & 28.0 & 44.0
& 14.0 & 26.0 & 12.0 & 80.0 & 31.6 \\

\rowcolor{cbstripe}
Qwen3.7-Plus
& 60.0 & 42.0 & 14.0 & 48.0 & 46.0
& 38.0 & 46.0 & 54.0 & 84.0 & 48.0 \\

Qwen3.8-27B
& 36.0 & 42.0 & 20.0 & 34.0 & 42.0
& 42.0 & 48.0 & 26.0 & 86.0 & 41.8 \\

\rowcolor{cbstripe}
Qwen3.8-Max
& 84.0 & 44.0 & \cbsecond{30.0} & 72.0 & 74.0
& 74.0 & 70.0 & 78.0 & 86.0 & 68.0 \\

Kimi-K2.6
& 30.0 & 42.0 & 14.0 & 38.0 & 46.0
& 48.0 & 40.0 & 38.0 & 86.0 & 42.4 \\

\rowcolor{cbstripe}
GPT-5.6-Luna
& 42.0 & \cbsecond{50.0} & 12.0 & 54.0 & 48.0
& 70.0 & 62.0 & 34.0 & 90.0 & 51.3 \\

GPT-5.6-Terra
& 78.0 & 40.0 & 20.0 & 36.0 & 26.0
& 62.0 & 58.0 & 60.0 & 90.0 & 52.2 \\

\rowcolor{cbstripe}
GPT-5.6-Sol
& \cbsecond{86.0}
& 46.0
& 28.0
& 64.0
& 54.0
& \cbsecond{82.0}
& \cbsecond{80.0}
& \cbsecond{88.0}
& \cbsecond{92.0}
& \cbsecond{68.9} \\

\midrule[0.6pt]

\rowcolor{cbheader}
\cbgroup{Chart-Specialized Models/Agents}

DePlot
& 18.0 & 34.0 & 8.0 & 4.0 & 6.0
& 2.0 & 2.0 & 4.0 & 0.0 & 8.7 \\

\rowcolor{cbstripe}
TinyChart
& 32.0 & 38.0 & 16.0 & 20.0 & 22.0
& 10.0 & 8.0 & 4.0 & 4.0 & 17.1 \\

ChartInstruct
& 20.0 & 16.0 & 6.0 & 8.0 & 4.0
& 8.0 & 6.0 & 4.0 & 6.0 & 8.7 \\

\rowcolor{cbstripe}
ChartVLM
& 16.0 & 30.0 & 6.0 & 8.0 & 6.0
& 2.0 & 4.0 & 6.0 & 8.0 & 9.6 \\

ChartAgent
& 30.0 & 20.0 & 6.0 & \cbsecond{94.0} & \cbsecond{80.0}
& 22.0 & 20.0 & 6.0 & 64.0 & 38.0 \\

\midrule[0.6pt]
\rowcolor{cbours}
\textcolor{cborange}{\textbf{ChartEvo}} (Ours)
& \cbbest{100.0}
& \cbbest{92.0}
& \cbbest{92.0}
& \cbbest{98.0}
& \cbbest{96.0}
& \cbbest{90.0}
& \cbbest{88.0}
& \cbbest{90.0}
& \cbbest{96.0}
& \cbbest{93.6} \\

\bottomrule[1.0pt]
\end{tabular}%
}
\endgroup
\vspace{-0.25cm}
\end{table*}

\vspace{-0.2cm}
\begin{table*}[th]
\centering
\caption{\textbf{Comparison on ChartX.}
NQA accuracy (\%) on eight long-tail chart types.
The best and second-best results are highlighted in
\textbf{bold} and \underline{underline}, respectively.}
\label{tab:chartx}

\begingroup
\scriptsize
\setlength{\tabcolsep}{4pt}
\renewcommand{\arraystretch}{1.18}

\definecolor{cxheader}{HTML}{F5F3F1}
\definecolor{cxstripe}{HTML}{FAF9F8}
\definecolor{cxours}{HTML}{FDEBDC}
\definecolor{cxorange}{HTML}{E96B25}
\definecolor{cxrule}{HTML}{BDB7B0}
\definecolor{cxgrouptext}{HTML}{655F59}

\arrayrulecolor{cxrule}

\newcommand{\cxbest}[1]{\textbf{#1}}
\newcommand{\cxsecond}[1]{\underline{#1}}

\newcommand{\cxgroup}[1]{%
  \multicolumn{10}{c}{%
    \color{cxgrouptext}\bfseries\itshape #1%
  }\\
}

\resizebox{\textwidth}{!}{%
\begin{tabular}{l|c*{7}{c}|c}
\toprule[1.0pt]

\rowcolor{cxheader}
\textbf{Method}
& \makebox[\widthof{\textbf{Radar(24)}}][c]{\textbf{Radar}}
& \makebox[\widthof{\textbf{3D-Bar(28)}}][c]{\textbf{3D-Bar}}
& \makebox[\widthof{\textbf{Rose(39)}}][c]{\textbf{Rose}}
& \makebox[\widthof{\textbf{Area(25)}}][c]{\textbf{Area}}
& \makebox[\widthof{\textbf{Bubble(32)}}][c]{\textbf{Bubble}}
& \makebox[\widthof{\textbf{Multi-axes(39)}}][c]{\textbf{Multi-axes}}
& \makebox[\widthof{\textbf{Treemap(41)}}][c]{\textbf{Treemap}}
& \makebox[\widthof{\textbf{Ring(37)}}][c]{\textbf{Ring}}
& \textbf{Avg.} \\
\midrule[0.6pt]

\rowcolor{cxheader}
\cxgroup{Direct Visual MLLMs}

DeepSeek-v4.1-flash
& 4.2 & 25.0 & 35.9 & 28.0 & 62.5
& 46.2 & 75.6 & 70.3 & 43.4 \\

\rowcolor{cxstripe}
Qwen3.7-Plus
& 16.7 & 21.4 & 46.2 & 48.0 & 65.6
& 79.5 & 75.6 & 75.7 & 53.6 \\

Qwen3.8-27B
& 8.3 & 14.3 & 43.6 & 40.0 & 62.5
& 56.4 & 75.6 & 67.6 & 46.0 \\

\rowcolor{cxstripe}
Qwen3.8-Max
& 54.2 & 32.1 & 79.5 & 84.0 & \cxbest{87.5}
& \cxsecond{89.7} & 87.8 & 67.6 & 72.8 \\

Kimi-K2.6
& 12.5 & 28.6 & 61.5 & 40.0 & 75.0
& 74.4 & 58.5 & 75.7 & 53.3 \\

\rowcolor{cxstripe}
GPT-5.6-Luna
& 41.7 & 32.1 & 59.0 & \cxsecond{88.0} & 78.1
& 84.6 & \cxsecond{90.2} & 81.1 & 69.4 \\

GPT-5.6-Terra
& 66.7 & \cxsecond{46.4} & 82.1 & \cxbest{92.0} & 71.9
& 84.6 & 85.4 & \cxsecond{94.6} & 78.0 \\

\rowcolor{cxstripe}
GPT-5.6-Sol
& \cxsecond{75.0}
& \cxsecond{46.4}
& \cxbest{97.4}
& \cxbest{92.0}
& \cxsecond{84.4}
& 87.2
& \cxbest{95.1}
& \cxbest{97.3}
& \cxsecond{84.4} \\

\midrule[0.6pt]

\rowcolor{cxheader}
\cxgroup{Chart-Specialized Models/Agents}

ChartInstruct
& 8.3 & 10.7 & 10.3 & 8.0 & 6.2
& 5.1 & 9.8 & 5.4 & 8.0 \\

\rowcolor{cxstripe}
ChartVLM
& 25.0 & 28.6 & 17.9 & 12.0 & 25.0
& 15.4 & 14.6 & 29.7 & 21.0 \\

DePlot
& 20.8 & 14.3 & 5.1 & 16.0 & 31.2
& 15.4 & 14.6 & 5.4 & 15.4 \\

\rowcolor{cxstripe}
TinyChart
& 25.0 & 28.6 & 15.4 & 24.0 & 25.0
& 23.1 & 7.3 & 10.8 & 19.9 \\

ChartAgent
& 29.2 & 28.6 & 28.2 & 32.0 & 68.8
& 15.4 & 65.9 & 51.4 & 39.9 \\
\midrule[0.6pt]
\rowcolor{cxours}
\textcolor{cxorange}{\textbf{ChartEvo}} (Ours)
& \cxbest{83.3}
& \cxbest{57.1}
& \cxsecond{89.7}
& \cxbest{92.0}
& \cxsecond{84.4}
& \cxbest{92.3}
& \cxbest{95.1}
& \cxsecond{94.6}
& \cxbest{86.1} \\

\bottomrule[1.0pt]
\end{tabular}%
}
\endgroup
\vspace{-0.5cm}

\end{table*}

\section{Experiments}
\vspace{-0.1cm}

We evaluate ChartEvo on two chart reasoning benchmarks, ChartBench and ChartX. We focus on Numerical Question Answering (NQA), which demands precise visual evidence extraction and numerical reasoning, providing a comprehensive evaluation of an agent's chart reasoning capabilities. We select 9 long-tail chart types from ChartBench and 8 from ChartX. For each chart type, we use 100 training questions for self-evolution and evaluate on a total of 450 test questions from ChartBench and 265 from ChartX. Self-evolution is conducted independently for the two benchmarks.

Following common evaluation practices in chart reasoning, we adopt a 5\% relative-error tolerance to account for uncertainty in visual representations and numerical readings. After self-evolution, the resulting harness is used for evaluation on the test set. For charts with explicit numerical annotations, the Agent directly performs visual answering.

We employ Qwen3.7-Plus as the main Agent and GPT-5.6-Sol as the agentic proposer. The maximum Agent action budget is set to 15. We compare ChartEvo against two groups of baselines: (1) direct MLLMs, including DeepSeek-v4.1-Flash~\cite{deepseekv41flash}, Qwen3.7-Plus~\cite{qwen37}, Qwen3.8-27B, Qwen3.8-Max~\cite{qwen3.8}, Kimi-K2.6, GPT-5.6-Luna, GPT-5.6-Terra, and GPT-5.6-Sol~\cite{gpt56}; and (2) chart-specialized methods, including DePlot~\cite{deplot}, TinyChart~\cite{tinychart}, ChartInstruct~\cite{chartinstruct}, ChartVLM~\cite{chartx}, and ChartAgent~\cite{chartagent}.

\begin{table}[t]
\centering
\caption{\textbf{Comparison of Inference Costs.}
Average inference cost per question across ChartBench and ChartX.
Lower values indicate lower cost.}
\label{tab:supp_cost}

\begingroup
\footnotesize
\setlength{\tabcolsep}{9pt}
\renewcommand{\arraystretch}{1.18}

\definecolor{cxheader}{HTML}{F5F3F1}
\definecolor{cxstripe}{HTML}{FAF9F8}
\definecolor{cxours}{HTML}{FDEBDC}
\definecolor{cxorange}{HTML}{E96B25}
\definecolor{cxrule}{HTML}{BDB7B0}

\arrayrulecolor{cxrule}

\begin{tabular}{lcc}
\toprule[1.0pt]

\rowcolor{cxheader}
\textbf{Method}
& \textbf{Tokens} $\downarrow$
& \textbf{Time (s)} $\downarrow$ \\
\midrule[0.6pt]

Qwen3.7-Plus
& 2,454.5 & 42.71 \\

\rowcolor{cxstripe}
Qwen3.8-27B
& 672.1 & 11.90 \\

Qwen3.8-Max
& 2,975.4 & 61.79 \\

\rowcolor{cxstripe}
Kimi-K2.6
& 3,612.1 & 46.70 \\

DeepSeek-v4.1-Flash
& 8,193.7 & 84.29 \\

\midrule[0.6pt]

\rowcolor{cxours}
\textcolor{cxorange}{\textbf{ChartEvo}} (Ours)
& 2,907.0 & 68.76 \\

\bottomrule[1.0pt]
\end{tabular}

\endgroup
\vspace{-0.4cm}
\end{table}

\vspace{-0.2cm}
\subsection{Comparison with State-of-the-art Solvers}
\vspace{-0.1cm}

Tables~\ref{tab:chartbench} and~\ref{tab:chartx} present the comparison of ChartEvo with existing methods on ChartBench and ChartX, respectively.
On ChartBench, ChartEvo achieves an average accuracy of 93.6\% across nine long-tail chart types, outperforming its base model, Qwen3.7-Plus, by 45.6 percentage points (93.6\% vs. 48.0\%), while surpassing all evaluated direct visual MLLMs and chart-specialized models and agents. Notably, ChartEvo achieves the best performance across all nine chart types, with eight reaching at least 90\% accuracy, demonstrating consistent improvements across diverse chart structures. For instance, on 3D-stack charts with complex spatial structures, ChartEvo improves accuracy from 14\% to 92\%.
On ChartX, ChartEvo also demonstrates substantial overall improvements across eight long-tail chart types, achieving an average accuracy of 86.1\% and slightly outperforming GPT-5.6-Sol. Improvements over the base model are observed across chart types with diverse visual structures and reasoning requirements. For example, on Radar charts, ChartEvo increases accuracy from 16.7\% to 83.3\%.
Overall, these results demonstrate that recursively evolving the external harness can substantially improve the capabilities of the underlying model in long-tail chart reasoning.

\vspace{-0.2cm}
\subsection{Cost Analysis}

We evaluate the inference cost of ChartEvo against direct visual MLLMs using completion tokens and wall-clock time. All results are averaged over the same 715 test questions from ChartBench and ChartX, providing a consistent basis for comparing inference overhead.

As shown in Table~\ref{tab:supp_cost}, ChartEvo requires an average of 2,907 completion tokens and 68.76 seconds per question, representing an 18\% increase in tokens and an additional 26.05 seconds compared with its base model, Qwen3.7-Plus. Its token consumption remains below Kimi-K2.6 and DeepSeek-v4.1-Flash, and is comparable to Qwen3.8-Max, while its latency is lower than DeepSeek-v4.1-Flash. The additional overhead is associated with multi-step Agent--Tool interactions, which enable structured evidence acquisition and numerical reasoning. Combined with the substantial accuracy improvements reported above, these results highlight the performance--cost trade-off of ChartEvo: stronger long-tail chart reasoning is achieved with moderate additional token generation, accompanied by increased execution latency.

\vspace{-0.2cm}
\subsection{Ablation Study}

\subsubsection{Ablation on Skill and Tool Evolution}
\vspace{-0.1cm}

\begin{table}[t]
\centering
\caption{\textbf{Ablation of Skill and Tool Evolution.}
NQA accuracy (\%) on five ChartBench chart types.}
\label{tab:skill_tool_ablation}

\begingroup
\footnotesize
\setlength{\tabcolsep}{12pt}
\renewcommand{\arraystretch}{1.18}

\definecolor{cxheader}{HTML}{F5F3F1}
\definecolor{cxstripe}{HTML}{FAF9F8}
\definecolor{cxours}{HTML}{FDEBDC}
\definecolor{cxorange}{HTML}{E96B25}
\definecolor{cxrule}{HTML}{BDB7B0}

\arrayrulecolor{cxrule}

\begin{tabular}{lc}
\toprule[1.0pt]

\rowcolor{cxheader}
\textbf{Configuration}
& \textbf{Accuracy (\%)} \\
\midrule[0.6pt]

Base MLLM (Qwen3.7-Plus)
& 42.8 \\

\rowcolor{cxstripe}
with Evolved Skill-only
& 55.6 \\

with Base Harness
& 57.2 \\

\rowcolor{cxstripe}
with Evolved Tools
& 84.4 \\

\midrule[0.6pt]

\rowcolor{cxours}
\textcolor{cxorange}{\textbf{ChartEvo}} (Full)
& \textbf{94.8} \\

\bottomrule[1.0pt]
\end{tabular}

\endgroup
\vspace{-0.3cm}
\end{table}
\vspace{-0.1cm}

To assess Skill and Tool evolution, we perform ablations on five ChartBench chart types: Area, 3D-stacked, Ring, Sector, and Radar multi-fill. We compare five configurations: Base MLLM (Qwen3.7-Plus), Evolved Skill-only, Base Harness, Evolved Tools, and full ChartEvo. Evolved Skill-only incorporates evolved reasoning guidance, Base Harness provides generic executable operations, and Evolved Tools-only includes evolved Tools without evolved Skills. As shown in Table~\ref{tab:skill_tool_ablation}, Evolved Skill-only and Base Harness achieve accuracies of 55.6\% and 57.2\%, respectively, both outperforming the Base MLLM at 42.8\%. Evolved Tools-only further reaches 84.4\%, a 27.2 percentage-point improvement over Base Tools.

\begin{figure}[h]
\vspace{-0.2cm}
    \centering
    \includegraphics[width=1.0\linewidth]{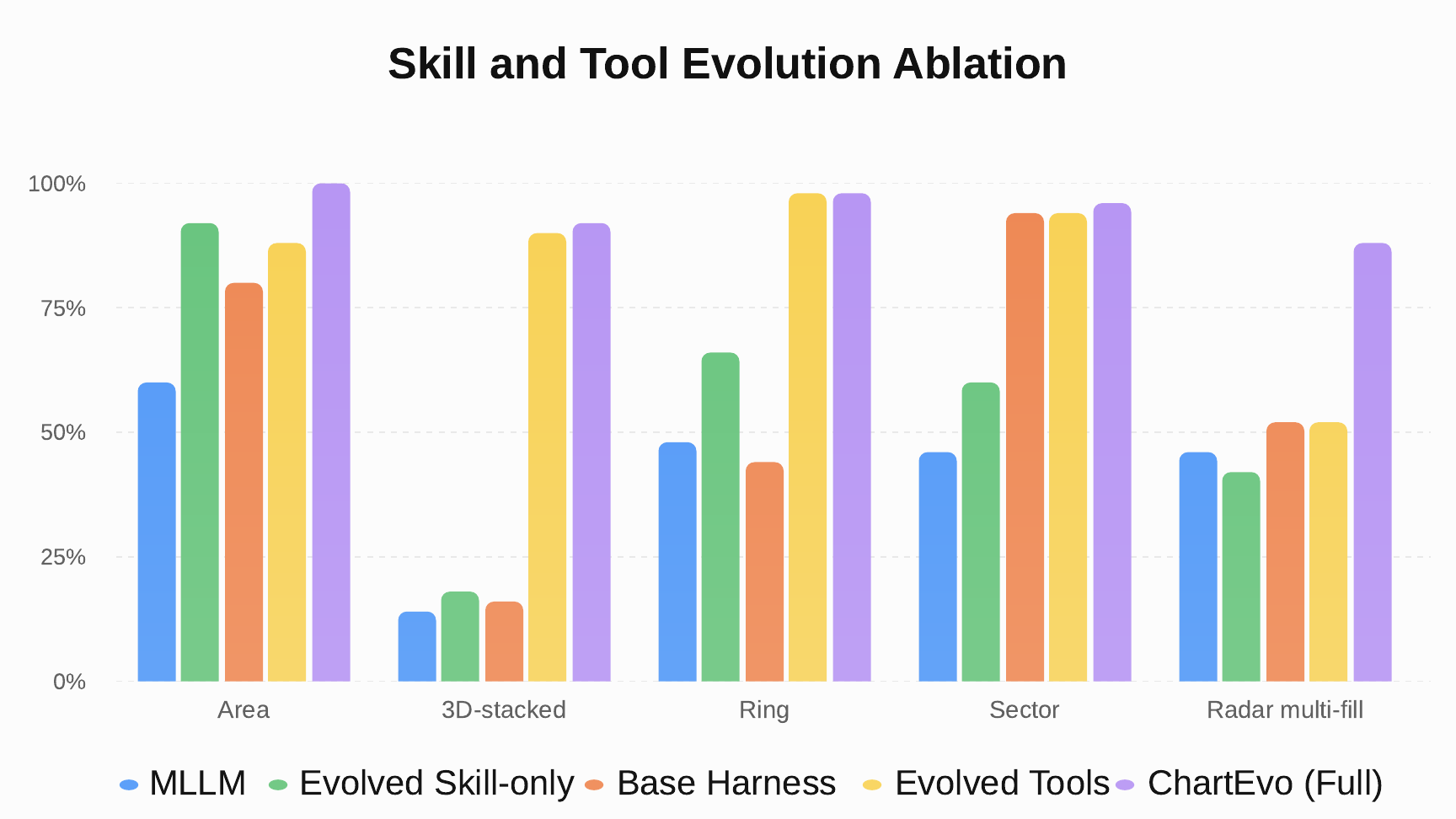}
    \caption{Per-type NQA accuracy on five ChartBench chart types.}
    \label{fig:ablation}
\vspace{-0.4cm}
\end{figure}

Despite similar overall accuracy, Skill-only and Base Tools exhibit distinct performance patterns, as shown in Fig.~\ref{fig:ablation}. On Area charts, Skill-only achieves 92\%, exceeding Base Tools at 80\%, demonstrating the value of chart-specific reasoning guidance. Conversely, Base Tools achieves 94\% on Sector charts, compared with 60\% for Skill-only, highlighting the importance of executable numerical measurements. These results reveal the complementary roles of Skills and Tools. Both configurations, however, remain below 20\% on 3D-stacked charts. With evolved Tools, accuracy increases from 16\% to 90\% on 3D-stacked and from 44\% to 98\% on Ring, primarily benefiting from specialized 3D measurement and ring-marker counting capabilities. Meanwhile, Sector maintains 94\% accuracy as its required measurements are already supported by existing Tools. Radar multi-fill also remains at 52\%, reflecting limitations in Tool coverage and effective invocation. These results indicate that Tool evolution depends on both acquiring the necessary executable capabilities and effectively utilizing them. The full ChartEvo further improves Radar multi-fill accuracy to 88\%, demonstrating the benefits of the integrated evolved harness.

\vspace{-0.1cm}
\subsubsection{Analysis of SELT}

\begin{table}[t]
\centering
\caption{\textbf{Inference Execution Statistics.}
Execution path distribution, accuracy, and inference
costs across ChartBench and ChartX.}
\label{tab:blc_analysis}

\begingroup
\footnotesize
\setlength{\tabcolsep}{6pt}
\renewcommand{\arraystretch}{1.18}

\definecolor{cxheader}{HTML}{F5F3F1}
\definecolor{cxstripe}{HTML}{FAF9F8}
\definecolor{cxrule}{HTML}{BDB7B0}

\arrayrulecolor{cxrule}

\begin{tabular}{lcc}
\toprule[1.0pt]

\rowcolor{cxheader}
\textbf{Execution Path}
& \textbf{Ratio (\%)}
& \textbf{Accuracy (\%)} \\
\midrule[0.6pt]

Chain-based Completion
& 77.9 & 98.0 \\

\rowcolor{cxstripe}
Direct Visual Answering
& 9.9 & 94.4 \\

Fallback
& 12.2 & 44.8 \\

\midrule[0.6pt]

\rowcolor{cxheader}
\multicolumn{3}{c}{\textbf{Average Calls per Question}} \\
\midrule[0.6pt]

Avg. API Calls
& \multicolumn{2}{c}{2.30} \\

\rowcolor{cxstripe}
Avg. Tool Calls
& \multicolumn{2}{c}{5.52} \\

\bottomrule[1.0pt]
\end{tabular}

\endgroup
\vspace{-0.4cm}
\end{table}

\begin{table}[t]
\centering
\caption{\textbf{SELT Ablation on ChartBench.}
Comparison of inference accuracy and execution costs
with and without persistent SELT across five chart types.}
\label{tab:selt_ablation}

\begingroup
\footnotesize
\setlength{\tabcolsep}{7pt}
\renewcommand{\arraystretch}{1.18}

\definecolor{cxheader}{HTML}{F5F3F1}
\definecolor{cxstripe}{HTML}{FAF9F8}
\definecolor{cxours}{HTML}{FDEBDC}
\definecolor{cxorange}{HTML}{E96B25}
\definecolor{cxrule}{HTML}{BDB7B0}

\arrayrulecolor{cxrule}

\begin{tabular}{lccc}
\toprule[1.0pt]

\rowcolor{cxheader}
\textbf{Configuration}
& \textbf{Acc. (\%)}
& \textbf{API Calls}
& \textbf{Tool Calls} \\
\midrule[0.6pt]

\rowcolor{cxstripe}
w/o SELT
& 91.2 & 6.19 & 5.41 \\

\midrule[0.6pt]

\rowcolor{cxours}
\textcolor{cxorange}{\textbf{ChartEvo}} (Full)
& \textbf{94.8}
& \textbf{2.73}
& \textbf{5.27} \\

\bottomrule[1.0pt]
\end{tabular}

\endgroup
\vspace{-0.4cm}
\end{table}

To isolate the contribution of Self-Evolving Local Toolchains (SELT), we conduct a controlled ablation on five ChartBench chart types: Area, 3D-stacked, Ring, Sector, and Radar multi-fill. Both configurations retain the same evolved Skills, Tools, and base model, while the variant without SELT relies on dynamic Tool composition instead of persistent Toolchains. As shown in Table~\ref{tab:selt_ablation}, enabling SELT reduces average model API calls from 6.19 to 2.73 (55.9\% reduction), while improving accuracy from 91.2\% to 94.8\%. This comparison evaluates whether reusing validated Tool compositions can reduce repeated model--Tool interactions without compromising reasoning performance.

We analyze the inference behavior of ChartEvo  across ChartBench and ChartX. As shown in Table~\ref{tab:blc_analysis}, 77.9\% of questions are completed through structured Tool chains, while 9.9\% are answered directly through visual inference and 12.2\% through fallback. Chain-based answers achieve 98.0\% accuracy, indicating the reliability of structured visual and numerical execution. ChartEvo averages 2.30 model API calls and 5.52 underlying Tool calls per question, demonstrating how multiple executable operations can be coordinated with limited model interactions. Together, the controlled ablation and execution statistics characterize the efficiency benefits of persistent SELT and the broad coverage of structured Tool-chain reasoning.

\vspace{-0.25cm}
\section{Conclusion}
\vspace{-0.2cm} 
This paper presents ChartEvo, a harness-level recursive self-improvement framework for long-tail chart reasoning. By evolving Skills, Tools, and Self-Evolving Local Toolchains, ChartEvo enables autonomous capability acquisition and efficient execution. On ChartBench and ChartX, it improves accuracy over Qwen3.7-Plus by 45.6\% and 32.5\%, respectively, while SELT substantially reduce model calls without accuracy loss. These results demonstrate the effectiveness of harness-level self-improvement for long-tail chart adaptation.

\section*{Limitations}
The effectiveness of ChartEvo depends partly on the coverage and diversity of the data used for self-evolution. Limited training examples may lead to overly specialized harness components, while conservative validation and admission criteria can restrict the acquisition of new capabilities. Furthermore, substantial distribution shifts between evolution and test data may limit the transferability of evolved Skills, Tools, and SELT. Improving the robustness and generalization of harness-level self-improvement remains an important direction for future work.

\bibliography{custom}

\clearpage
\appendix
\section{Chart Question Answering Problem Formulation}
Chart Question Answering (CQA) is formulated as a multimodal reasoning task in which a system is given a chart image $\mathcal{I}_i$ and a corresponding natural language question $\mathcal{Q}_i$, and is required to produce an answer $\mathcal{A}_i$. In this work, we focus on Numerical Question Answering (NQA), where the answer is a numerical value that must be inferred from the visual and quantitative information encoded in the chart. Solving these questions may require identifying relevant chart elements, extracting numerical evidence, and performing quantitative reasoning to derive the final answer.

\section{Experimental Details}
\subsection{Benchmark Selection and Data Organization}
We evaluate ChartEvo on two chart question-answering benchmarks, ChartBench and ChartX, both of which cover diverse chart types and provide numerical question-answer pairs. We focus on Numerical Question Answering (NQA) and select chart types where the base model exhibits relatively weak performance for subsequent self-evolution and evaluation.

Specifically, we first evaluate Qwen3.7-Plus on the randomly selected 100 questions from train splits of ChartBench and ChartX to obtain per-type NQA accuracy. Chart types with a baseline accuracy below 85\% are selected as long-tail chart types. This procedure yields 9 chart types from ChartBench and 8 from ChartX.

For each selected chart type, we use 100 training questions to drive harness-level self-evolution and evaluate the evolved system on the corresponding test questions. Self-evolution is conducted independently for ChartBench and ChartX, with separate evolving harnesses for the two benchmarks. The resulting evaluation sets contain 450 NQA questions across 9 chart types in ChartBench and 265 NQA questions across 8 chart types in ChartX.

\subsection{Self-Evolution Implementation}
\subsubsection{Training Organization.}
For each long-tail chart type, we partition 100 training questions into 10 clusters of 10 questions each. Self-evolution proceeds sequentially across clusters, with the current harness used to execute the tasks in each cluster. The resulting answers, correctness feedback, and execution trajectories are collected to guide subsequent analysis and evolution.
Within each cluster, harness evolution may involve multiple iterations. After each update, the system re-evaluates relevant tasks to assess whether previous failures have been resolved and continues refining the harness to address remaining failures. Once most failures within the cluster have been addressed, the process advances to the next cluster. Verified and accepted harness updates are persisted, allowing subsequent clusters to build upon the accumulated Skills, Tools, and SELT.

\subsubsection{Failure Analysis and Harness Updates}
\textbf{Failure Identification.}
After completing each training cluster, the system identifies tasks requiring further investigation based on final answers, correctness feedback, and execution trajectories. For NQA tasks, predictions with a relative error exceeding 5\% are marked as incorrect, while ground-truth values of zero are handled separately to avoid division by zero. Additionally, predictions with errors between 3\% and 5\% are flagged for further inspection, as numerically correct answers may still arise from unreliable execution trajectories. Beyond answer correctness, the system also examines fallback behavior and the task coverage of existing Tool-call chains. Together, these signals identify tasks that are not yet reliably handled by the current harness and guide subsequent evolution.

\textbf{Failure Clustering and Trajectory Analysis.}
The agentic proposer analyzes the collected execution trajectories to identify recurring failure patterns across training questions. Rather than relying solely on numerical discrepancies, the analysis considers the Agent's action sequences, Tool invocations, and returned observations to locate the underlying sources of failure. These may include incorrect Tool selection, parameter binding, or evidence utilization despite sufficient existing capabilities; missing Tools for extracting necessary visual or numerical evidence; or ineffective execution order and information dependencies across multiple Tool calls. By clustering failures with shared characteristics, the proposer identifies recurring capability gaps that can guide reusable harness improvements across tasks.

\textbf{Harness Updates and Verification.}
Based on the identified failure patterns, the agentic proposer proposes targeted harness updates by revising existing Skills, Tools, and SELT or introducing new components. A single evolution iteration may jointly update multiple component types to coordinate capability acquisition, capability utilization, and execution structure. Candidate updates are evaluated by re-executing relevant training tasks with the modified harness and examining both answer correctness and the resulting execution trajectories. Verification considers whether previous failures have been resolved, whether the Agent correctly invokes the updated components, whether Tool executions produce useful evidence, and whether tasks previously unsupported by harness-based reasoning become solvable. Accepted updates are persisted, while remaining failures may trigger further rounds of analysis and evolution within the same training cluster.

\subsubsection{Harness Growth Through Self-Evolution}
We further examine how the ChartEvo harness evolves across
ChartBench and ChartX. As shown in
Table~\ref{tab:harness_growth}, both benchmarks start from
the same initial harness containing four Skills, 25 available
Tools, and no SELT. After self-evolution, the number of
distinct Skills increases to 13 on ChartBench and 12 on ChartX,
while the number of available Tools expands to 42 and 60,
respectively. Here, available Tools are counted as the union
of Tool names permitted by all Skills in the final harness,
rather than the number of candidate Tool versions generated
during evolution.

ChartEvo also accumulates 20 active SELT on ChartBench
and 71 on ChartX, indicating that self-evolution extends
beyond individual Skills and Tools to reusable execution
structures. ChartBench retains slightly more distinct Skills,
whereas ChartX exhibits greater Tool expansion and
substantially more active SELT. These results illustrate
how ChartEvo progressively accumulates and organizes
reusable capabilities within its external harness.
\begin{table}[h]
\centering
\caption{\textbf{Harness Growth Through Self-Evolution.}
Comparison of harness components before and after
self-evolution on ChartBench and ChartX.}
\label{tab:harness_growth}

\begingroup
\footnotesize
\setlength{\tabcolsep}{6pt}
\renewcommand{\arraystretch}{1.18}

\definecolor{cxheader}{HTML}{F5F3F1}
\definecolor{cxstripe}{HTML}{FAF9F8}
\definecolor{cxours}{HTML}{FDEBDC}
\definecolor{cxorange}{HTML}{E96B25}
\definecolor{cxrule}{HTML}{BDB7B0}

\arrayrulecolor{cxrule}

\begin{tabular}{lccc}
\toprule[1.0pt]

\rowcolor{cxheader}
\textbf{Component}
& \textbf{Initial}
& \textbf{ChartBench}
& \textbf{ChartX} \\
\midrule[0.6pt]

Distinct Skills
& 4 & 13 (+9) & 12 (+8) \\

\rowcolor{cxstripe}
Available Tools
& 25 & 42 (+17) & 60 (+35) \\

\rowcolor{cxours}
Active SELT
& 0 & 20 & 71 \\

\bottomrule[1.0pt]
\end{tabular}

\endgroup
\vspace{-0.4cm}
\end{table}

\subsubsection{Illustrative Self-Evolution Example}
\begin{figure*}[th]
  \centering
  \includegraphics[width=\linewidth]{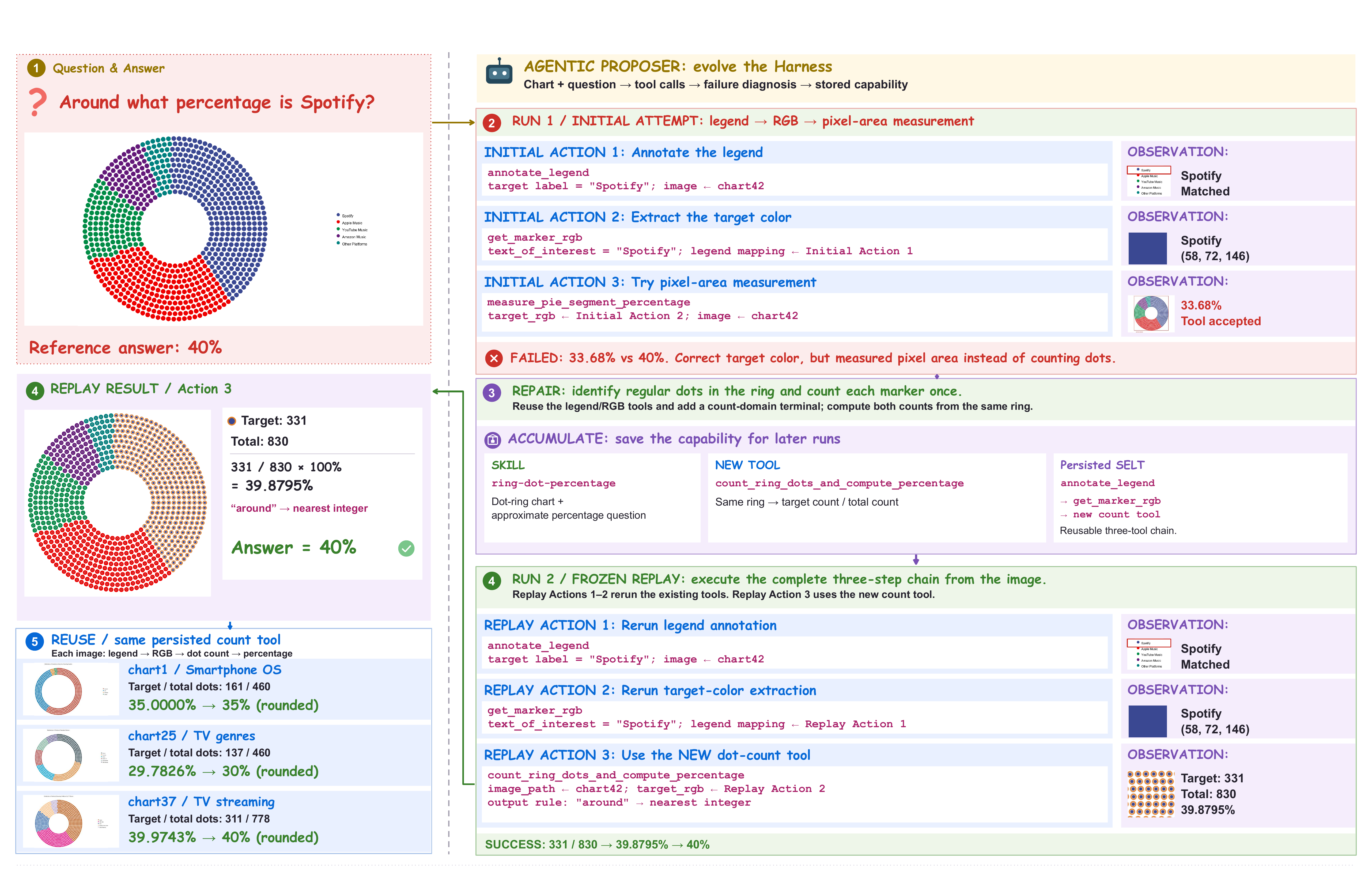}
  \vspace{-0.8cm}
  \caption{Illustrative self-evolution of ChartEvo on a dot-ring chart. An initial measurement failure triggers joint Skill, Tool, and SELT evolution, enabling accurate dot-count-based reasoning and capability reuse across additional chart instances.}
  \label{fig:supp_evolution_example}
\vspace{-0.5cm}
\end{figure*}

Figure~\ref{fig:supp_evolution_example} illustrates how ChartEvo acquires reusable harness capabilities through self-evolution using a dot-ring chart question asking for the approximate percentage of Spotify, with a reference answer of 40\%. In the initial execution, the Agent first invokes \texttt{annotate\_legend} to identify the target series, followed by \texttt{get\_marker\_rgb} to extract its corresponding color. It then applies \texttt{measure\_pie\_segment\_percentage} to estimate the target proportion from pixel area, obtaining 33.68\%. Although the target series and its color are correctly identified and the measurement Tool returns an \texttt{accepted} status, the final prediction deviates from the reference answer. Trajectory analysis reveals a mismatch between the measurement strategy and the chart's visual encoding: proportions are represented by discrete dot counts rather than pixel area.

To address this failure, ChartEvo jointly evolves multiple harness components. A new Tool, \texttt{count\_ring\_dots\_and\_compute\_percentage}, is introduced to count target-colored dots and total dots within the same ring and compute their ratio. A corresponding Skill, \texttt{ring-dot-percentage}, provides capability selection and usage guidance for dot-ring charts and approximate percentage questions. Meanwhile, the existing legend annotation and color extraction Tools are composed with the new counting Tool into a three-step SELT, preserving the complete measurement procedure as a reusable execution structure. These coordinated updates extend the harness with a measurement capability tailored to discrete-dot encoding, together with the knowledge and execution structure required to utilize it.

When executing the original question with the evolved harness, the Agent follows the three-step Tool chain to identify the target series, extract its color, and count the corresponding dots. The new Tool identifies 331 target dots out of 830 total dots, yielding 39.8795\%, which is rounded to 40\% in accordance with the question's approximate wording. The same persisted counting capability is further reused on three additional dot-ring charts involving smartphone operating systems, TV genres, and TV streaming, producing answers of 35\%, 30\%, and 40\%, respectively. This example illustrates how ChartEvo transforms an observed measurement failure into a persistent harness capability that can be applied across subsequent chart instances.

\subsection{Ablation Study Configurations}
All ablation settings use Qwen3.7-Plus as the base model and are evaluated on the same test questions under a unified 5\% relative-error tolerance. Specifically, \textit{Base MLLM} performs direct visual answering without any harness components. \textit{Evolved Skill-only} uses evolved Skills to guide reasoning without access to Tools or SELT. \textit{Base Harness} includes the initial Skills and Tools, without SELT. \textit{Evolved Tools} retains the initial Skills while replacing the initial Tool collection with its evolved counterpart, again without SELT. Finally, \textit{ChartEvo (Full)} employs the complete evolved harness, comprising evolved Skills, the same evolved Tools, and persisted SELT.

\section{Harness Specifications}
\subsection{Harness Representation and Organization}
ChartEvo organizes the capabilities acquired through self-evolution into a persistent harness comprising Skills, Tools, and Self-Evolving Local Toolchains (SELT). Each component adopts a representation suited to its role. Skills are stored as metadata-annotated textual knowledge, specifying applicability conditions, reasoning guidance, and knowledge for capability selection and utilization. Tools are registered as executable components with explicit input and output interfaces for visual and numerical processing. SELT preserve validated Tool-call compositions as reusable execution structures with defined functional contracts, execution order, and internal dependencies. Together, these components form an extensible collection of capabilities accessible to the Agent during reasoning.

Harness components are connected through explicit references that coordinate knowledge, executable capabilities, and execution structures. Skills provide reasoning guidance for particular chart structures and question types, together with the selection and usage knowledge of relevant Tools and the registration and routing information of SELT. Guided by these Skills, the Agent can invoke individual Tools, dynamically compose multiple Tool calls, or reuse persisted SELT. Each SELT specifies its constituent Tool calls, execution order, and parameter dependencies across steps, enabling multiple executable capabilities to operate within a predefined structure. These cross-component references allow reasoning knowledge, executable capabilities, and execution structures to be organized and utilized together.

Verified and accepted harness updates are persisted in the corresponding asset stores and remain available throughout subsequent training stages and inference tasks. Existing components may be revised to accommodate new capability requirements, while new Skills, Tools, or SELT can be introduced to extend the range of supported tasks and execution patterns. When an evolution iteration involves multiple component types, their associated guidance, executable capabilities, and execution structures can be updated jointly and retained as interconnected assets. This persistent organization enables the harness to accumulate reusable capabilities across successive evolution iterations and supports subsequent reasoning with the expanded capability repertoire.

\subsection{Hierarchical Retrieval and Progressive Disclosure}
\label{sec:supp_harness_retrieval}

As the harness evolves, its growing collection of Skills, Tools, and SELT introduces an increasing amount of potentially relevant knowledge and executable capabilities. ChartEvo employs hierarchical retrieval and progressive disclosure to construct a task-relevant execution context, selectively exposing harness information according to task characteristics and evolving execution states.

ChartEvo adopts a two-stage retrieval mechanism to select relevant Skills without loading the entire Skill Store. In the first stage, Skill frontmatter metadata is used to filter and rank candidates based on chart type, question characteristics, and applicability conditions. Only the selected Skills are subsequently loaded for reasoning. For each question, the system retrieves at most three Skills, comprising up to one universal Skill and two chart-specific Skills, with a combined Skill-body budget of 1,600 tokens. Near-duplicate Skills with similarity exceeding 0.80 are suppressed, while chart- or feature-specific matches are prioritized over overly general or mismatched candidates. This mechanism limits the knowledge introduced into the Agent's context while preserving guidance relevant to the current reasoning task.

Beyond initial retrieval, ChartEvo progressively discloses Skill content according to observed execution states. Each Skill may contain general routing guidance alongside conditional instructions associated with different Tool outcomes. At the beginning of a task, the Runtime exposes only the initial routing guidance rather than all possible execution and recovery branches. As reasoning proceeds, Tool outcomes such as \texttt{accepted}, \texttt{rejected}, and \texttt{argument\_error} trigger the disclosure of corresponding Skill branches. Instructions associated with unobserved conditions remain hidden. Consequently, a Skill can encode normal execution, error handling, and recovery guidance while exposing only the instructions relevant to the current stage of reasoning.

ChartEvo organizes executable capabilities through a two-level Tool Catalog. The overview level provides the names and concise functional descriptions of all registered Tools, allowing the Agent to discover available capabilities. The detail level selectively exposes complete interfaces for a small set of relevant Tools, including input schemas, output fields, invocation constraints, and usage examples. During initial context construction, at most six detailed Tool interfaces are disclosed; after context refresh, this limit is reduced to four. Candidate selection incorporates signals from the retrieved Skills, Tools referenced by relevant SELT, question semantics, recently invoked Tools, and Tools associated with preceding execution failures. This separation preserves broad capability discoverability while limiting the context occupied by detailed executable interfaces.

The set of available SELT is further restricted by the retrieved Skills and task-specific applicability conditions. The Runtime first collects SELT references from the selected Skills and then filters candidates according to chart metadata, question semantics, and parameter bindability. An eligible SELT must satisfy its chart applicability conditions, match relevant question terms, and admit the binding of required selectors from the question, legend, or axis metadata. Candidates with unresolved references, ambiguous identifiers, or unsatisfied execution prerequisites are excluded from the available SELT catalog. The remaining SELT are presented to the Agent as reusable execution structures, from which the Agent may select an appropriate composition for the current task.

Together, these mechanisms enable ChartEvo to organize an expanding harness into a compact, task-relevant reasoning context. Skill knowledge, executable Tool interfaces, and reusable SELT are disclosed selectively, with subsequent guidance adapting to observed execution states.

\subsection{Harness Execution and Runtime Contracts}
Following harness retrieval and disclosure, the Agent selects individual Tools, reuses existing SELT, or dynamically composes Tool calls based on the current question, chart information, and accumulated observations. For individual Tools, the Agent constructs invocation arguments using relevant chart elements, data series, axis information, and available evidence. For SELT, the Agent determines whether the functional contract matches the current task and binds the required external parameters, while the internal execution order and parameter dependencies are predefined by the persisted composition. This separation allows the Agent to retain control over semantic decisions and capability selection while delegating predefined execution procedures to the Runtime.

The Runtime executes the selected harness capabilities and validates their outputs against the corresponding contracts. For individual Tools, it performs the requested operations using the Agent-provided arguments and returns structured visual or numerical evidence. For SELT, the Runtime executes the constituent Tool calls sequentially according to the predefined composition, resolves parameter dependencies across steps, and aggregates the resulting observations. These internal Tool calls proceed without requiring a separate Agent decision at each step, reducing repeated MLLM--Tool interactions. Both individual Tools and SELT are subject to contract validation, including input/output interface constraints and component-specific functional requirements. Successful execution alone does not guarantee that the resulting evidence is accepted; the outputs must also satisfy the corresponding validity conditions before being used in subsequent reasoning. This contract-aware execution mechanism provides the Agent with structured evidence accompanied by explicit acceptance states.

After execution, the Runtime returns the resulting observations and execution states to the Agent. Outputs satisfying the applicable contracts are marked as \texttt{accepted}, allowing the Agent to incorporate the evidence into subsequent numerical reasoning or Tool invocations. Outputs that fail the relevant acceptance conditions are marked as \texttt{rejected}, while invalid invocation arguments produce states such as \texttt{argument\_error}. Based on this feedback, the Agent may revise arguments, select alternative Tools, reorganize subsequent execution steps, or proceed with the available evidence. These execution states also inform the conditional Skill disclosure mechanism described in Section~\ref{sec:supp_harness_retrieval}, providing additional guidance aligned with the observed execution outcomes. Through this interaction, semantic reasoning and contract-aware execution remain coordinated throughout multi-step chart reasoning.

\subsection{Initial Harness Configuration}
To clarify the initial capability boundary of ChartEvo, we describe the harness configuration used before self-evolution. Both ChartBench and ChartX start from the same initial harness and undergo independent self-evolution. The initial harness consists of 4 Skills and 25 generic Tools, providing foundational capabilities for chart understanding, visual evidence extraction, and numerical processing. No SELT are predefined in the initial harness. The complete inventories of initial Skills and Tools, together with their functional descriptions, are provided in Tables~\ref{tab:initial_skills} and~\ref{tab:initial_tools}.

\begin{table*}[t]
\centering
\caption{Initial Skill inventory of ChartEvo before self-evolution, summarizing the reasoning and capability-use guidance provided by each Skill.}
\label{tab:initial_skills}
\small
\setlength{\tabcolsep}{7pt}
\renewcommand{\arraystretch}{1.12}
\begin{tabular}{lp{12cm}}
\toprule
\textbf{Skill} & \textbf{Reasoning and Capability-Use Guidance} \\
\midrule
\texttt{chart-evidence-validation}
& Reuse validated observations, distinguish rejected from advisory evidence, preserve numerical semantics, and avoid ambiguous bindings or invalid observation references. \\
\midrule
\texttt{bar-value}
& Guide legend and category binding, axis calibration, bar localization, and numerical projection, with handling of stacked segments and negative bars. \\
\midrule
\texttt{line-value}
& Guide series localization, point extraction, and axis projection, while addressing error-bar anchors, ambiguous candidates, and coordinate consistency. \\
\midrule
\texttt{pie-percentage}
& Guide color and label binding, segment measurement, and percentage estimation using geometric and area-based evidence. \\
\bottomrule
\end{tabular}
\vspace{-0.2cm}
\end{table*}

\begin{table*}[ht]
\centering
\caption{Initial Tool inventory of ChartEvo before self-evolution. The 25 generic Tools are grouped by their primary functionalities.}
\label{tab:initial_tools}
\small
\setlength{\tabcolsep}{7pt}
\renewcommand{\arraystretch}{1.08}
\begin{tabular}{lp{11.5cm}}
\toprule
\textbf{Tool} & \textbf{Functionality} \\
\midrule
\multicolumn{2}{l}{\textit{OCR, Text, Color, and Segmentation (7)}} \\
\midrule
\texttt{locate\_text} & Locate visible text labels and return their bounding boxes and positions. \\
\texttt{annotate\_legend} & Identify specified legend labels and construct their spatial mappings. \\
\texttt{get\_marker\_rgb} & Extract the RGB color of a marker associated with a legend label. \\
\texttt{get\_embedded\_label\_rgb} & Extract the underlying fill color associated with an embedded data label. \\
\texttt{clean\_chart\_image} & Remove detected textual regions to produce a cleaned chart image. \\
\texttt{segment\_and\_mark} & Generate candidate segmentation masks and indexed visual overlays. \\
\texttt{resolve\_text\_to\_mask} & Associate a target text label with a candidate segmentation mask. \\
\midrule
\multicolumn{2}{l}{\textit{Axis, Coordinates, and Deterministic Computation (5)}} \\
\midrule
\texttt{axis\_tick\_localizer} & Locate specified axis tick labels and their pixel positions. \\
\texttt{axis\_localizer} & Estimate numerical axis calibration from detected tick information. \\
\texttt{interpolate\_pixel\_to\_value} & Convert pixel coordinates into numerical values using reference anchors. \\
\texttt{project\_point\_to\_axis\_value} & Project a selected point onto a calibrated numerical axis. \\
\texttt{arithmetic} & Perform deterministic arithmetic operations on supplied operands. \\
\midrule
\multicolumn{2}{l}{\textit{Pie and Area Measurement (3)}} \\
\midrule
\texttt{measure\_pie\_segment\_percentage} & Estimate pie or donut segment proportions from color-based area measurements. \\
\texttt{compute\_segment\_area} & Measure selected region areas and compute proportions with a specified denominator. \\
\texttt{measure\_mask\_angular\_percentage} & Estimate angular proportions of resolved radial segmentation masks. \\
\midrule
\multicolumn{2}{l}{\textit{Bar and Boxplot (4)}} \\
\midrule
\texttt{get\_bar} & Locate axis-aligned bars and extract their bounding boxes. \\
\texttt{compute\_bar\_height} & Convert detected bar geometry into numerical values using axis anchors. \\
\texttt{get\_boxplot} & Extract geometric landmarks of axis-aligned boxplots. \\
\texttt{compute\_boxplot\_entity} & Calculate boxplot statistics from extracted landmarks and axis references. \\
\midrule
\multicolumn{2}{l}{\textit{Line, Area, and Scatter (3)}} \\
\midrule
\texttt{get\_edgepoints} & Extract local edge points associated with a target color. \\
\texttt{locate\_series\_point} & Locate candidate series points near a specified horizontal position. \\
\texttt{locate\_color\_extent} & Measure color extents along a specified scanline. \\
\midrule
\multicolumn{2}{l}{\textit{Radial Geometry (3)}} \\
\midrule
\texttt{get\_radial} & Extract radial contours, centers, and outer radii. \\
\texttt{analyze\_radial\_geometry} & Analyze radial geometry and estimate aggregate radius references. \\
\texttt{estimate\_radial\_value} & Estimate numerical values from radii and supplied reference values. \\
\bottomrule
\end{tabular}
\vspace{-0.2cm}
\end{table*}

\section{Case Study}
We present qualitative analyses of ChartEvo's inference trajectories to examine how the evolved harness supports numerical chart reasoning in practice. We organize the cases into three categories: successful reasoning, error recovery, and unsuccessful reasoning. Together, they illustrate the utilization of structured evidence, adaptation to intermediate execution feedback, and remaining limitations in challenging chart scenarios.

\vspace{-0.1cm}
\subsection{Successful Reasoning Cases}
We first examine successful reasoning trajectories to illustrate how ChartEvo utilizes its evolved harness to solve challenging numerical questions. These cases highlight the coordination of multiple executable capabilities, including target identification, visual evidence extraction, and numerical calibration, and how their intermediate observations contribute to the final prediction.

\begin{figure*}[th]
  \centering
  \includegraphics[width=\linewidth]{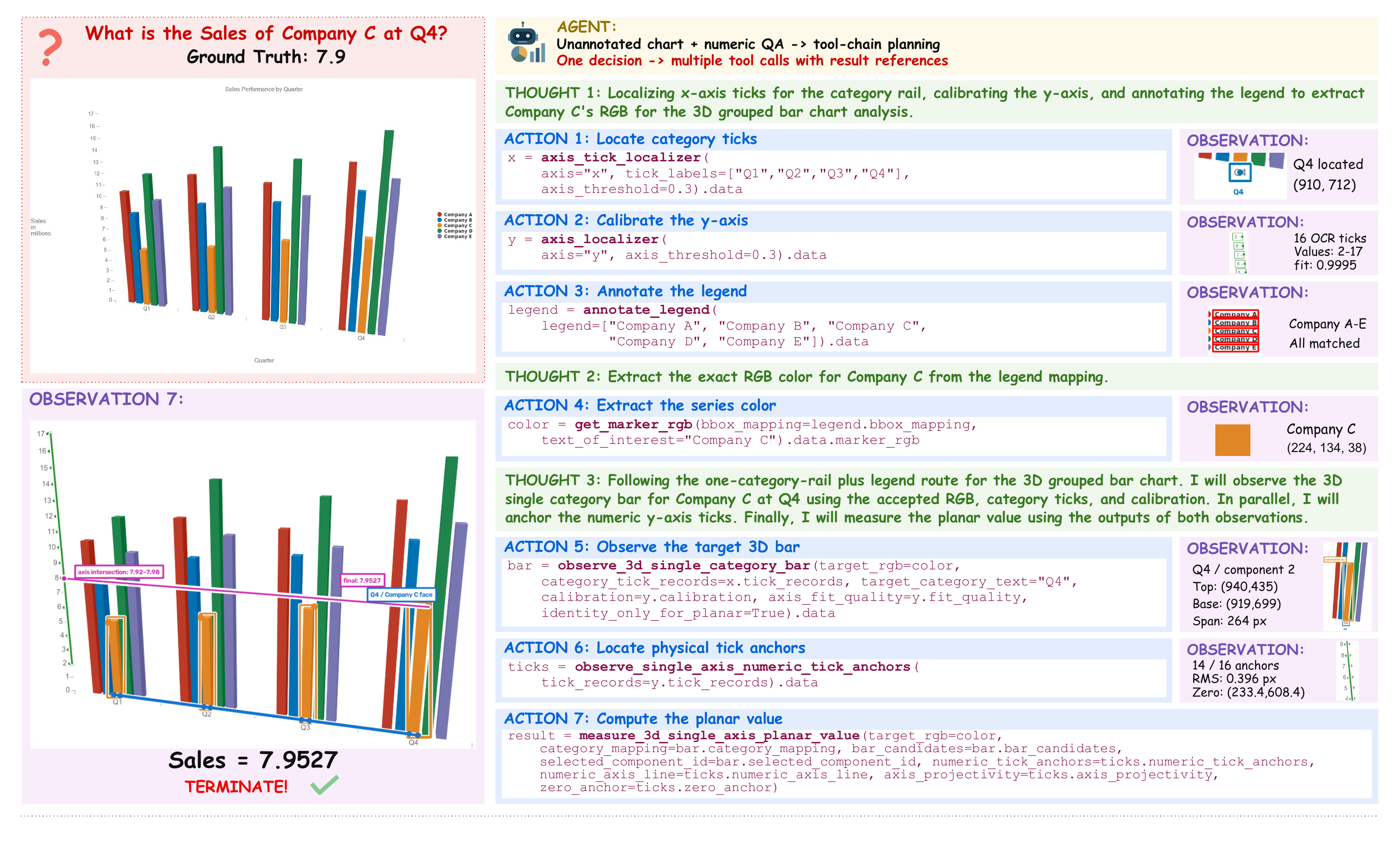}
  \vspace{-0.8cm}
  \caption{Successful reasoning on a 3D grouped bar chart. ChartEvo coordinates multiple Tool calls to extract structured evidence and accurately estimate the target value.}
  \label{fig:case_success_1}
\vspace{-0.5cm}
\end{figure*}

Figure~\ref{fig:case_success_1} presents an example involving a 3D grouped bar chart. ChartEvo identifies the target category and series, calibrates the numerical axis, and integrates geometric observations with axis anchors to estimate the target value. The resulting prediction of 7.9527 closely matches the ground truth of 7.9, illustrating the effectiveness of structured evidence integration for numerical reasoning over complex chart geometry.

\vspace{-0.1cm}
\subsection{Error Recovery Cases}
We further investigate reasoning trajectories in which ChartEvo encounters intermediate execution errors but ultimately reaches correct predictions. These cases highlight how the Agent interprets Tool feedback, revises its actions, and recovers from unsuccessful intermediate steps.

\begin{figure*}[th]
  \centering
  \includegraphics[width=\linewidth]{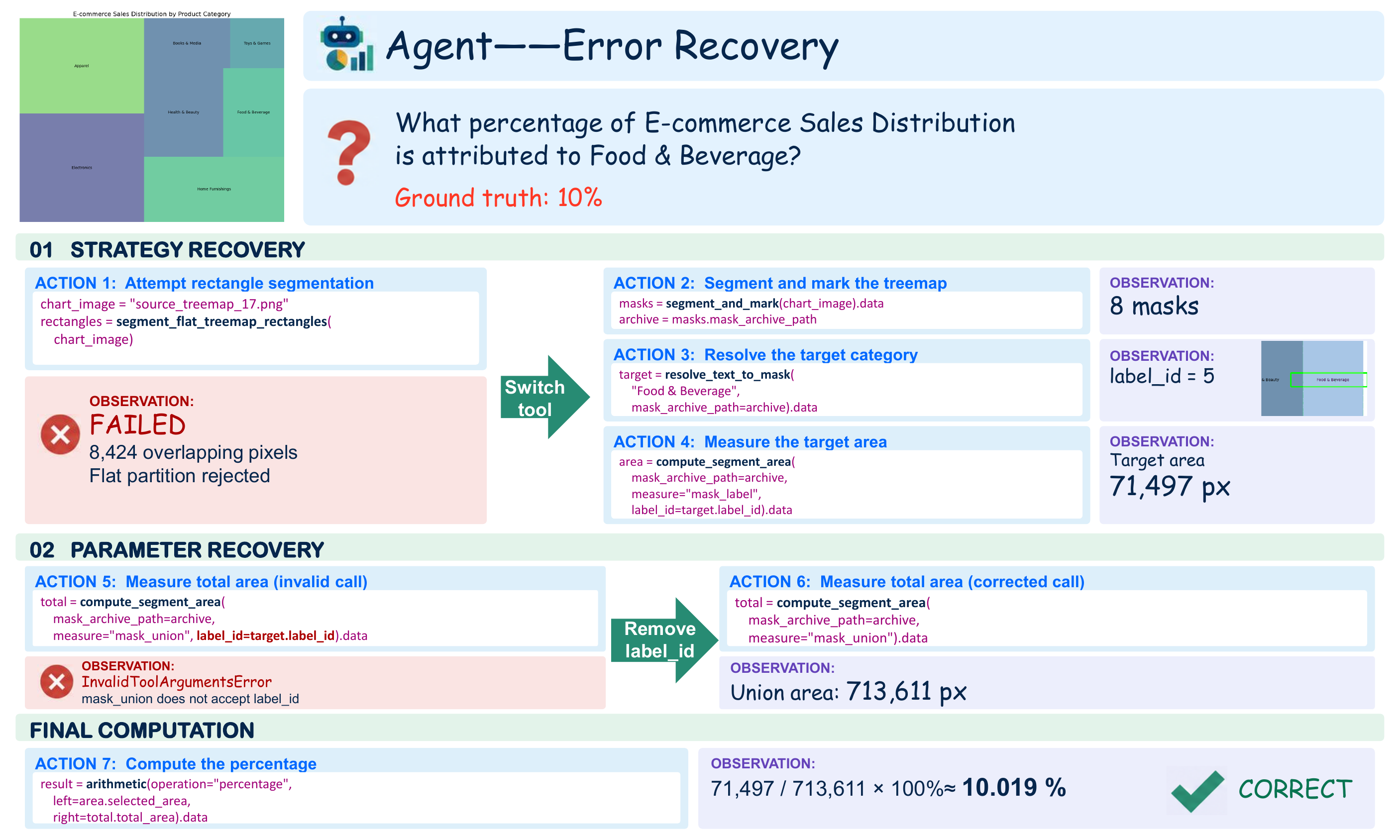}
  \vspace{-0.8cm}
  \caption{Error recovery on a treemap. ChartEvo switches segmentation strategies after an initial failure and corrects an invalid Tool argument, ultimately producing an accurate percentage estimate.}
  \label{fig:case_recovery}
\vspace{-0.5cm}
\end{figure*}

Figure~\ref{fig:case_recovery} illustrates two forms of recovery within a single treemap reasoning trajectory. The Agent initially attempts rectangle-based segmentation, which fails due to overlapping regions. It then switches to mask-based segmentation, successfully identifying the target category and measuring its area. A subsequent attempt to measure the total area triggers an argument error because an incompatible parameter is supplied. After removing the invalid argument, the Agent obtains the total area and computes a percentage of 10.019\%, closely matching the ground truth of 10\%. This case demonstrates how execution feedback supports both strategy-level adaptation and parameter-level correction without requiring harness updates.

\vspace{-0.1cm}
\subsection{Failure Cases}
Finally, we examine unsuccessful reasoning trajectories to characterize the remaining limitations of ChartEvo. These cases investigate situations where the available harness capabilities or intermediate observations are insufficient for reliable numerical inference, as well as cases where the Agent fails to appropriately handle uncertain evidence.

\begin{figure*}[th]
  \centering
  \includegraphics[width=\linewidth]{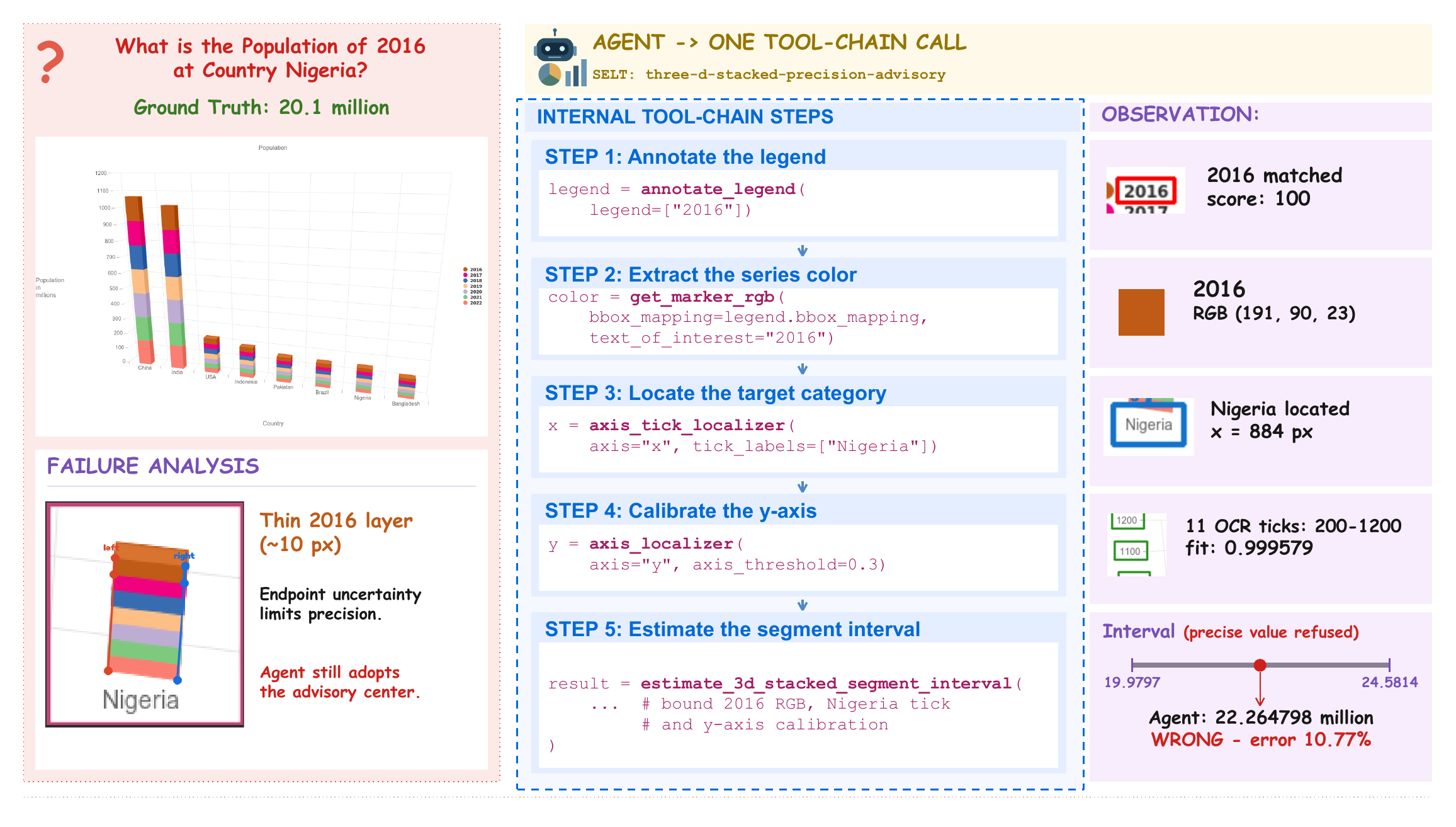}
  \vspace{-0.8cm}
  \caption{Failure case on a 3D stacked bar chart. Uncertain segment boundaries prevent precise measurement, while the Agent adopts an unreliable estimate, leading to an incorrect answer.}
  \label{fig:case_failure_1}
\vspace{-0.5cm}
\end{figure*}

Figure~\ref{fig:case_failure_1} illustrates a failure on a 3D stacked bar chart. Although the target series and category are correctly identified, the thin target segment introduces substantial uncertainty in endpoint localization. The measurement Tool consequently returns an estimated interval rather than an accepted precise value. However, the Agent adopts the interval midpoint, producing 22.2648 million against the ground truth of 20.1 million. This case highlights a remaining limitation in reasoning with uncertain numerical evidence.

\clearpage
\end{document}